\documentclass[]{article}

\usepackage{graphicx}

\usepackage{amsmath}
\usepackage{amsfonts}
\usepackage{amssymb}
\usepackage{url}
\usepackage{bm}
\usepackage{float}
\usepackage{algorithm}
\usepackage{algpseudocode}
\usepackage{multirow}
\usepackage{bbm}

\makeatletter
\newcommand{\algmargin}{\the\ALG@thistlm}
\makeatother
\newlength{\whilewidth}
\algdef{SE}[parWHILE]{parWhile}{EndparWhile}[1]
{\parbox[t]{\dimexpr\linewidth-\algmargin}{%
		\hangindent\whilewidth\strut\algorithmicwhile\ #1\ \algorithmicdo\strut}}{\algorithmicend\ \algorithmicwhile}%
\algnewcommand{\parState}[1]{\State%
	\parbox[t]{\dimexpr\linewidth-\algmargin}{\strut #1\strut}}

\DeclareMathOperator*{\argmax}{argmax}

\title{Listening while Speaking: \\Speech Chain by Deep Learning}

\author{Andros Tjandra, Sakriani Sakti, Satoshi Nakamura\\
	Graduate School of Information Science\\
	Nara Institute of Science and Technology, Japan \\
	\texttt{\{andros.tjandra.ai6,ssakti,s-nakamura\}@is.naist.jp}
}

\date{}

\begin{document}
	%\ninept
	%
	\maketitle
	\begin{abstract}
		Despite the close relationship between speech perception and production, research in automatic speech recognition (ASR) and text-to-speech synthesis (TTS) has progressed more or less independently without exerting much mutual influence on each other. In human communication, on the other hand, a closed-loop speech chain mechanism with auditory feedback from the speaker's mouth to her ear is crucial. In this paper, we take a step further and develop a closed-loop speech chain model based on deep learning. The sequence-to-sequence model in close-loop architecture allows us to train our model on the concatenation of both labeled and unlabeled data. While ASR transcribes the unlabeled speech features, TTS attempts to reconstruct the original speech waveform based on the text from ASR. In the opposite direction, ASR also attempts to reconstruct the original text transcription given the synthesized speech. To the best of our knowledge, this is the first deep learning model that integrates human speech perception and production behaviors. Our experimental results show that the proposed approach significantly improved the performance more than separate systems that were only trained with labeled data.
	\end{abstract}
	\section{Introduction}
	\label{sec:intro}
	
	The speech chain, which was first introduced by Denes et al. \cite{denes1993speech}, describes the basic mechanism involved in speech communication when a spoken message travels from the speaker’s mind to the listener’s mind (Fig.~\ref{fig:speech_chain}). It consists of a speech production mechanism in which the speaker produces words and generates speech sound waves, transmits the speech waveform through a medium (i.e., air), and creates a speech perception process in a listener’s auditory system to perceive what was said. Over the past few decades, tremendous research effort has struggled to understand the principles underlying natural speech communication. Many attempts have also been made to replicate human speech perception and production by machines to support natural modality in human-machine interactions.
	
	To date, the development of advanced spoken language technologies based on ASR and TTS has enabled machines to process and respond to basic human speech. Various ASR approaches have relied on acoustic-phonetics knowledge \cite{davis1952autodigit} in early works to template-based schemes with dynamic time warping (DTW) \cite{vintsyuk1968speechdynamic,sakoe1978dynamicprog} and data-driven approaches with rigorous statistical modeling of a hidden Markov model-Gaussian mixture model (HMM-GMM) \cite{jelinek1976hmm,wilpon1990hmm}. In a similar direction, TTS technology development has gradually shifted from the foundation of a rule-based system using waveform coding and an analysis-synthesis method \cite{olive1977rule,sagisaka1988rule} to a waveform unit concatenation approach \cite{sagisaka1992atr,hunt1996unit} and a more flexible approach using the statistical modeling of hidden semi-Markov model-GMM (HSMM-GMM) \cite{yoshimura1999hmm,tokuda1995hmm}. Recently, after the resurgence of deep learning, interest has also surfaced in the possibility of utilizing a neural approach for ASR and TTS systems. Many state-of-the-art performances in ASR \cite{graves2013speech,palaz2015convolutional,sainath2015learning} and TTS \cite{heiga2013tts,oord2016wavenet,arik2017deepvoice} tasks have been successfully constructed based on neural network frameworks.
	
	\begin{figure}[]
		\label{fig:speech_chain}
		\centering
		\includegraphics[width=\linewidth]{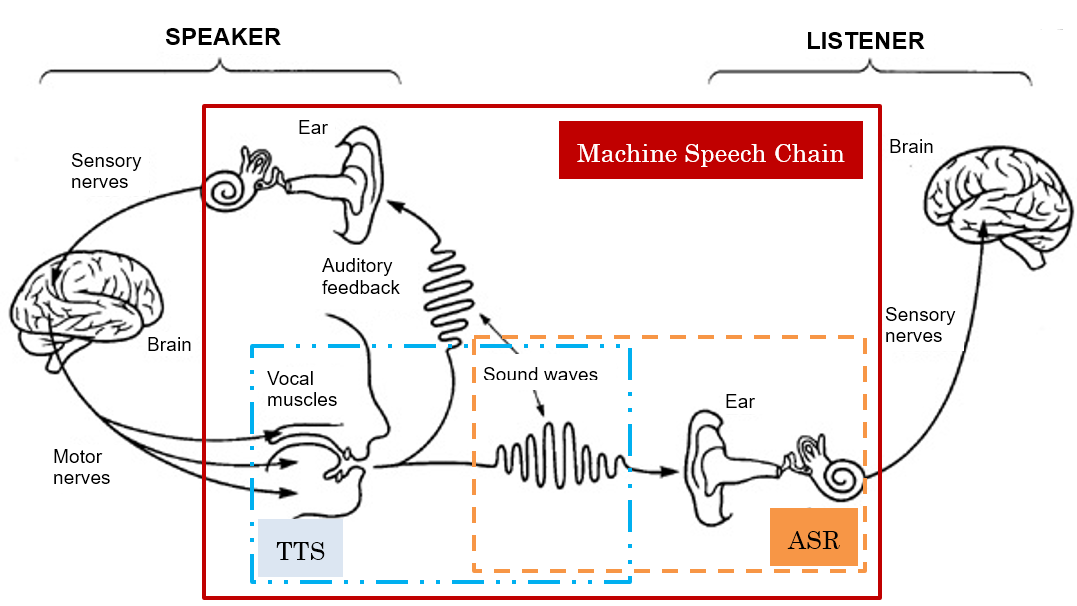}
		\caption{Speech chain \cite{denes1993speech} and related spoken language technologies.}
		\label{fig:speech_chain}
		
	\end{figure}
	
	\begin{figure*}[]
		\centering
		\includegraphics[width=0.87\linewidth]{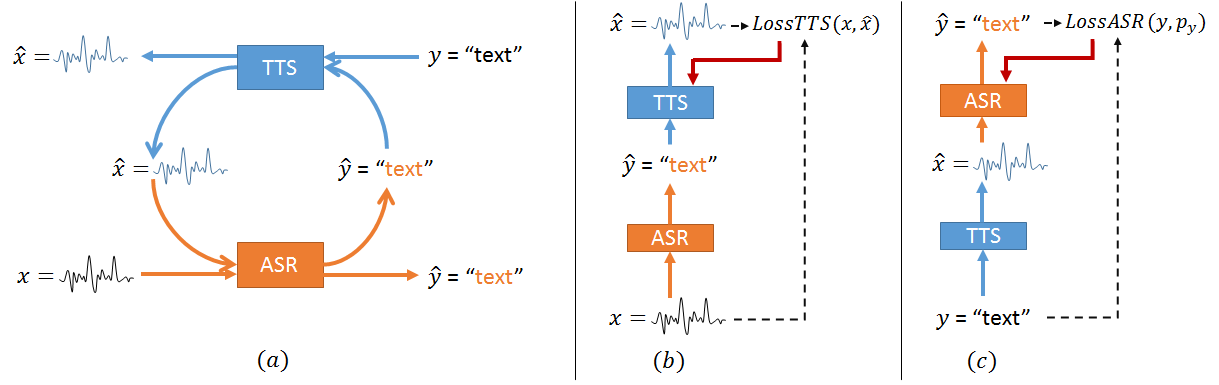}
		
		\caption{(a) Overview of machine speech chain architecture. Examples of unrolled process: (b) from ASR to TTS and (c) from TTS to ASR.}
		
		\label{fig:machine_chain}
	\end{figure*}
	
	However, despite the close relationship between speech perception and production, ASR and TTS researches have progressed more or less independently without exerting much mutual influence on each other. In human communication, on the other hand, a closed-loop speech chain mechanism has
	a critical auditory feedback mechanism from the speaker’s mouth to her ear (Fig.~\ref{fig:speech_chain}). In other words, the hearing
	process is critical, not only for the listener but also for the speaker. By simultaneously listening and speaking, the speaker can monitor her volume, articulation, and the general comprehensibility of her speech. Processing the information further, the speaker’s brain can plan what she will say next. Children who lose their hearing often have difficulty to produce clear speech due to inability to monitor their own speech.
	
	Unfortunately, investigating the inherent links between these two processes is very challenging. Difficulties arise because methodologies and analysis are necessarily quite different when they are extracting the underlying messages from speech waveforms as in speech perception or generating
	an optimum dynamic speaking style from the intended message as in speech production. Until recently, it was impossible in a joint approach to reunite  the problems shared by both modes. However, due to deep learning’s representational power, many complicated hand-engineered models have been simplified by letting DNNs learn their way from input to output spaces. With this newly emerging approach to sequence-to-sequence mapping tasks, a model with a common architecture can directly learn the mapping between variable-length representations of different modalities:
	text-to-text sequences \cite{bahdanau2014neural, sutskever2014sequence}, speech-to-text sequences \cite{chorowski2014end, chan2016listen}, text-to-speech sequences \cite{wang2017tacotron}, and image-to-text sequences \cite{kelvinxu2016image2caption}, etc.
	
	Therefore, in this paper, we take a step further and develop a closed-loop speech chain model based on deep learning and construct a sequence-to-sequence model for both ASR and TTS tasks as well as a loop connection between these two processes. The sequence-to-sequence model in closed-loop architecture allows us to train our model on the concatenation of both labeled and unlabeled data. While ASR transcribes the unlabeled
	speech features, TTS reconstructs the original speech waveform based on text from ASR. In the opposite direction, ASR also reconstructs the original text transcription given the synthesized speech. To the best of our knowledge, this is the first deep learning model that integrates human speech perception and production behaviors.

	\begin{algorithm}
		\footnotesize
		\caption{Speech Chain Algorithm}
		\label{alg:speech_chain}
		\begin{algorithmic}[1]
			\State \textbf{Input}:Paired speech and text dataset $\mathcal{D}^P$, text-only dataset $\mathcal{Y}^U$, speech-only dataset $\mathcal{X}^U$, supervised loss coefficient $\alpha$, unsupervised loss coefficient $\beta$
			\vspace{0.2cm}
			\Repeat
			\vspace{0.2cm}
			\State \textbf{A. Supervised training with speech-text data pairs}
			\vspace{0.1cm}
			\parState {%
				Sample paired speech and text\\ $(x^P, y^P) = ([x_1^P,..,x_{S_P}^P], [y_1^P,..,y_{T_P}^P])$ \\from $\mathcal{D}^P$ with speech length $S_P$ and text length $T_P$.}
			
			\parState{%
				Generate a text probability vector by ASR using teacher-forcing:\\
				$p_{y_t}=P_{ASR}(\cdot|y^P_{< t}, x^P; \theta_{ASR}), \forall t\in[1..T_P]$
			}
			\parState{%
				Generate best predicted speech by TTS using teacher-forcing: \\
				$\hat{x}_s^P = \argmax\limits_{z} P_{TTS}(z|x^P_{<s}, y^P; \theta_{TTS}); \forall s\in[1..S_P]$
			}
			\State Calculate the loss for ASR and TTS
			\vspace{-0.2cm}
			\begin{eqnarray}
			L_P^{ASR} = LossASR(y^P, p_{y}; \theta_{ASR}) \\
			L_P^{TTS} = LossTTS(x^P, \hat{x}^P; \theta_{TTS})
			\end{eqnarray}
			
			\vspace{0.1cm}
			\State \textbf{B. Unsupervised  training with unpaired speech and text}
			\vspace{0.1cm}
			\State \textbf{\textit{\# Unrolled process from TTS to ASR:}}
			\State Sample text $y^U = [y_1^U,..,y_{T_U}^U]$ from $\mathcal{Y}^{U}$
			\State Generate speech by TTS: $\hat{x}^U\sim P_{TTS}(\cdot | y^U; \theta_{TTS})$
			\parState{%
				Generate text probability vector by ASR from TTS's predicted speech using teacher-forcing:\\
				$p_{y_t}=P_{ASR}(\cdot|y^U_{< t}, \hat{x}^U; \theta_{ASR}), \forall t\in[1..T_U]$
			}
			\parState {%
				Calculate the loss between original text $y^U$ and reconstruction probability vector $p_{y_t}$}
			\vspace{-0.3cm}
			\begin{eqnarray}
			L_U^{ASR} &=& LossASR(y^U, p_{y}; \theta_{ASR})
			\end{eqnarray}
			\vspace{0.2cm}
			\State \textbf{\textit{\# Unrolled process from ASR to TTS:}}
			\State Sample speech $x^U=[x_1^U,..,x_{S_U}^U]$ from $\mathcal{X}^{U}$
			\State Generate text by ASR: $\hat{y}^U \sim P_{ASR}(\cdot | x^U; \theta_{ASR})$
			\parState {%
				Generate speech by TTS from ASR's predicted text using teacher-forcing:\\
				$\hat{x}_s^U = \argmax\limits_{z} P_{TTS}(z|x_{<s}^U, \hat{y}^U; \theta_{TTS}); \forall s\in[1..S]$}
			\parState {%
				Calculate the loss between original speech $x^U$ and generated speech $\hat{x}^U$}
			\vspace{-0.35cm}
			\begin{eqnarray}
			L_U^{TTS} &=& LossTTS(x^U, \hat{x}^U; \theta_{TTS})
			\end{eqnarray}
			\vspace{0.2cm}
			\State \textbf{\textit{\# Loss combination:}}

			\State Combine all weighted loss into a single loss variable
			\vspace{-0.1cm}
			\begin{eqnarray}
			L = \alpha * (L_P^{TTS}+L_P^{ASR}) + \beta   * (L_{U}^{TTS} + L_{U}^{ASR})
			\end{eqnarray}
			\vspace{-0.3cm}
			\parState {%
				Calculate TTS and ASR parameters gradient with \\
				the derivative of $L$ w.r.t $\theta_{ASR}, \theta_{TTS}$}
			\vspace{-0.35cm}
			\begin{eqnarray}
			G_{ASR} &=& \nabla_{\theta_{ASR}}{L} \\
			G_{TTS} &=& \nabla_{\theta_{TTS}}{L}
			\end{eqnarray}
			\vspace{-0.3cm}
			\parState {%
				Update TTS and ASR parameters with gradient descent \\
				optimization (SGD, Adam, etc)}
			\vspace{-0.3cm}
			\begin{eqnarray}
			\theta_{ASR} \leftarrow Optim(\theta_{ASR}, G_{ASR})\\
			\theta_{TTS} \leftarrow Optim(\theta_{TTS}, G_{TTS})
			\end{eqnarray}

			\Until{convergence of parameter $\theta_{TTS}, \theta_{ASR}$}
		\end{algorithmic}
	\end{algorithm}

	\section{Machine Speech Chain}
	 
	An overview of our proposed machine speech chain architecture is illustrated in Fig.~\ref{fig:machine_chain}(a). It consists of a sequence-to-sequence ASR, a sequence-to-sequence TTS, and a loop connection from ASR to TTS and from TTS to ASR. The key idea is to jointly train both the ASR and TTS models. As mentioned above, the sequence-to-sequence model in closed-loop architecture allows us to train our model on the concatenation of both the labeled and unlabeled data. For supervised training with labeled data (speech-text pair data), both models can be trained independently by minimizing the loss between their predicted target sequence and the ground truth sequence. However, for unsupervised training with unlabeled data (speech only or text only), both models need to support each other through a connection.
	
	To further clarify the learning process during unsupervised training, we unrolled the architecture as follows: 
	 
	\begin{itemize}
		\item \textbf{Unrolled process from ASR to TTS} \\
		Given the unlabeled speech features, ASR transcribes the unlabeled input speech, while TTS reconstructs the original speech waveform based on the output text from ASR. Fig.~\ref{fig:machine_chain}(b) illustrates the mechanism. We may also treat it as an autoencoder model, where the speech-to-text ASR serves as an encoder and the text-to-speech TTS as a decoder. 
		\item \textbf{Unrolled process from TTS to ASR} \\
		Given only the text input, TTS generates speech waveform, while ASR also reconstructs the original text transcription given the synthesized speech. Fig.~\ref{fig:machine_chain}(c) illustrates the mechanism. Here, we may also treat it as another autoencoder model, where the text-to-speech TTS serves as an encoder and the speech-to-text ASR as a decoder.
	\end{itemize}  
	 
	With such autoencoder models, ASR and TTS are able to teach each other by adding a reconstruction term of the observed unlabeled data to the training objective. Details of the algorithm can be found in Alg.~\ref{alg:speech_chain}.

	\section{Sequence-to-Sequence Model for ASR}
	 
\begin{figure}[H]
		\centering
		\includegraphics[width=0.70\linewidth]{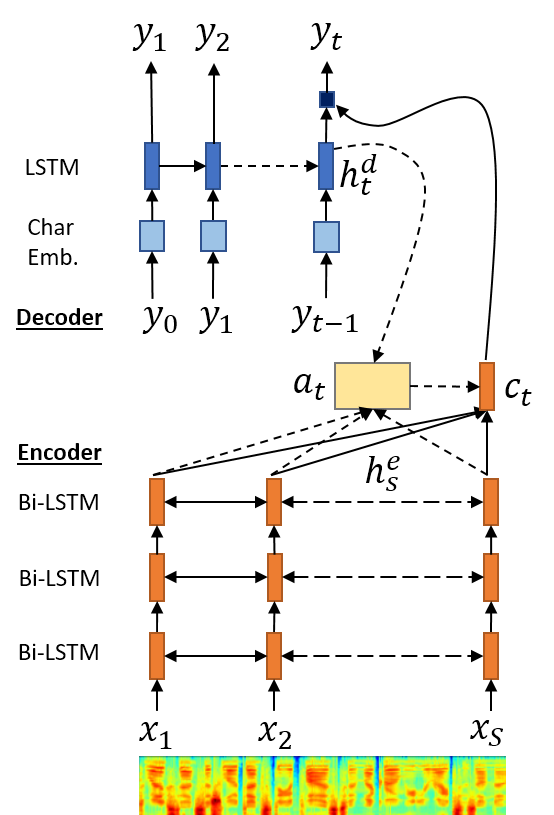}
		\caption{Sequence-to-sequence ASR architecture.}
		
		\label{fig:seq2seq_asr}
	\end{figure}

	A sequence-to-sequence model is a neural network that directly models conditional probability $p({y}|{x})$, where ${x} = [x_1, ..., x_S]$ is the sequence of the (framed) speech features with length $S$ and ${y} = [y_1, ..., y_T]$ is the sequence of label with length $T$. Fig.~\ref{fig:seq2seq_asr} shows the overall structure of the attention-based encoder-decoder model that consists of encoder, decoder and attention modules.

	The encoder task processes input sequence ${x}$ and outputs representative information ${h^e} = [h^e_1, ...,h^e_S]$ for the decoder. The attention module is an extension scheme that assist the decoder to find relevant information on the encoder side based on the current decoder hidden states \cite{bahdanau2014neural, chan2016listen}. Attention modules produces context information $c_t$ at time $t$ based on the encoder and decoder hidden states:
	\begin{align}
	c_t &= \sum_{s=1}^{S} a_t(s) * h^e_s \\
	a_t(s) &= \text{Align}({h^e_s}, h^d_t) \nonumber \\
	&= \frac{\exp(\text{Score}(h^e_s, h^d_t))}{\sum_{s=1}^{S}\exp(\text{Score}(h^e_s, h^d_t))}
	\end{align}
	There are several variations for score functions \cite{luong2015effective}:
	\begin{align}
	\text{Score}(h_s^e, h_t^d) =
	\begin{cases}
	\langle h_s^e, h_t^d\rangle, & \text{dot product}  \\
	h_s^{e\intercal} W_{s} h_t^d, & \text{bilinear}  \\
	V_s^{\intercal} \tanh(W_{s} [h_s^e, h_t^d]), & \text{MLP} \label{eq:mlpscore}  \\
	\end{cases}
	\end{align} where $\text{Score}:(\mathbb{R}^M \times \mathbb{R}^N) \rightarrow \mathbb{R}$, $M$ is the number of hidden units for the encoder and $N$ is the number of hidden units for the decoder.
	Finally, the decoder task predicts target sequence probability $p_{y_t}$ at time $t$ based on previous output and context information $c_t$. The loss function for ASR can be formulated as:
	\begin{equation}
	Loss_{ASR}(y, p_{y}) = -\frac{1}{T}\sum_{t=1}^{T}\sum_{c=1}^{C}\mathbbm{1}(y_t=c)*\log{p_{y_t}}[c]
	\end{equation} where $C$ is the number of output classes.
	Input ${x}$ for speech recognition tasks is a sequence of feature vectors like log Mel-scale spectrogram. Therefore, ${x} \in \mathbb{R}^{S \times D}$ where D is the number of features and S is the total frame length for an utterance. Output ${y}$, which is a speech transcription sequence, can be either phoneme or grapheme (character) sequence.
	 
	\section{Sequence-to-Sequence Model for TTS}
	 Parametric speech synthesis resembles a sequence-to-sequence task where we generate speech given a sentence. Using a sequence-to-sequence model, we model the conditional probability between $p(x|y)$, where $y=[y_1,...,y_T]$ is the sequence of characters with length $T$ and $x=[x_1,...,x_S]$ is the sequence of (framed) speech features with length $S$. From the sequence-to-sequence ASR model perspective, now we have an inverse model for reconstructing the original speech given the text.
	
	\begin{figure}[H]
		\centering
		\includegraphics[width=0.75\linewidth]{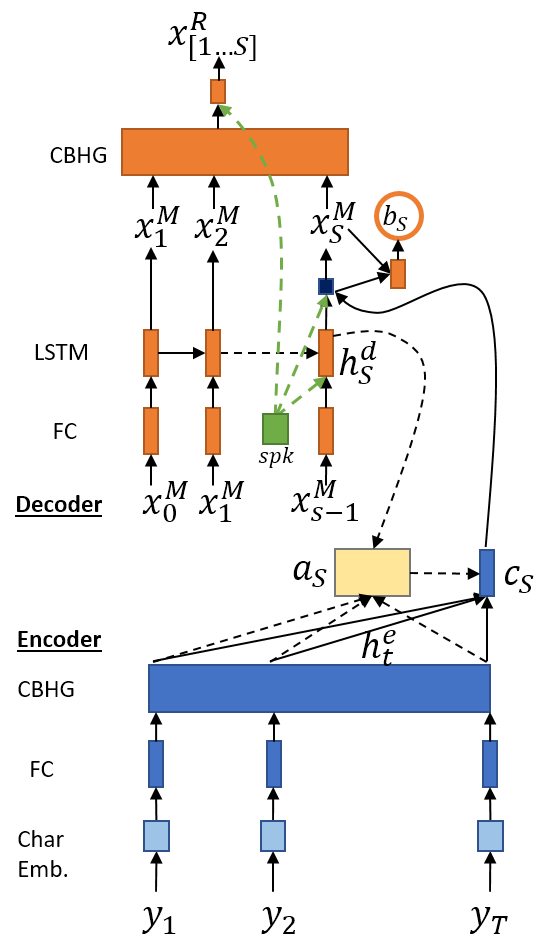}
		
		\caption{Sequence-to-sequence TTS (Tacotron) architecture with frame ending binary prediction. (FC = Fully Connected, CBHG = Convolution Bank + Highway + bi-GRU)}
		\label{fig:seq2seq_tts}
	\end{figure}

	In this work, our core architecture is based on Tacotron \cite{wang2017tacotron} with several structural modifications. Fig.~\ref{fig:seq2seq_tts} illustrates our modified Tacotron. In the encoder sides, we project our input characters with an embedding layer. The character vectors are fed into several fully connected layers followed by a non-linear activation function. We pass the result into the CBHG block (1-D \textbf{C}onvolution \textbf{B}ank + \textbf{H}ighway + bidirectional \textbf{G}RU) with eight filter banks (filter size from 1 to 8). The CBHG output is expected to produce representative information ${h^e} = [h_1^e,...,h_T^e]$ for the decoder.
	
	Our modified decoder has one input layer and three output layers (instead of two as in the original Tacotron). The first output layer generates log Mel-scale spectrogram $x^M = [x_1^M,...,x_S^M]$. At the $s$-th step, the input layer is fed by a previous step-log Mel-scale spectrogram $x_{s-1}^M$ and then several fully connected layers and a non-linear activation function are processed. Next we use a stacked LSTM with a monotonic attention mechanism \cite{graves2013generating} to extract expected context $c_s$ information based on the current decoder input and encoder states ${h^e}$. We project the context with a fully connected layer to predict the log Mel-scale spectrogram.
	
	The second output layer reconstructs log magnitude spectrogram $x^R = [x_1^R,...,x_S^R]$ given the first layer generated output $x^M$. After we get complete sequences  of the log Mel-scale spectrogram, we feed them into a CBHG block, followed by a fully connected layer to predict the log magnitude spectrogram.
	
	The third output layer generates binary prediction $b_s \in [0, 1]$ (1 if the $s$-th frame is the end of speech, otherwise 0) based on the current log-Mel spectrogram generated by the first output layer and expected context $c_s$ from the decoder with the attention layer. We add the binary prediction layer because the output from the first and second decoder layers is a real value vector, and we cannot use an end-of-sentence (eos) token  to determine when to stop the generation process. Based on our initial experiment, we found that our modification helped Tacotron determine the end of speech more robustly than forcing the decoder to generate frames with an 0 value at the end of the speech. We also enable our model to learn from multiple speaker by concatenating the projected speaker embedding into input before LSTM layer, first output regression layer, and second output regression layer.
	
	For training TTS model, we used the following loss function:
	\begin{equation}
	\begin{split}
	Loss_{TTS}(x, \hat{x}, b, \hat{b}) = & \frac{1}{S}\sum_{s=1}^{S} (x_s^M - \hat{x}_s^M)^2 + (x_s^R - \hat{x}_s^R)^2 \\
	& - (b_s \log(\hat{b}_s) + (1-b_s) \log(1-\hat{b}_s))
	\end{split}
	\end{equation} where $\hat{x}^M, \hat{x}^R, \hat{b}$ are  the predicted log Mel-scale spectrogram, the log magnitude spectrogram and the end-of-frame probability, and $x^M, x^R, b$ is the ground truth. In the decoding process, we use Griffin-Lim \cite{griffin1984signal} algorithm to iteratively estimate the phase spectrogram and reconstruct the signal with inverse STFT.
	 
	\section{Experiment on Single-Speaker Task}
	 
	To verify our proposed method, first we experimented on a corpus with a single speaker because until recently, most TTS systems by deep learning are trained on a single speaker dataset.

	To gather a large single speaker speech dataset, we utilized Google TTS \footnote{Google TTS  \url{https://pypi.python.org/pypi/gTTS}} to generate a large set of speech waveform based on basic travel expression corpus (BTEC) \cite{kikui2003creating} English sentences.
	For training and development we used part of the BTEC1 dataset, and for testing we used the default BTEC test set. For supervised training on both the ASR and TTS models, we chose 10,000 speech utterances that were paired with their corresponding text. For our development set, we selected another 3000 speech utterances and paired them with corresponding text. For our test set, we used all 510 utterances from the BTEC default test set. For the unsupervised learning step, we chose 40,000 speech  utterances just from BTEC1 and 40,000 text utterances from BTEC1. None of these sets overlap to each other.

	\subsection{Features Extraction}
	 
	For the speech features, we used a log magnitude spectrogram extracted by short-time Fourier transform (STFT) from the Librosa library \cite{librosa}. First, we applied wave-normalization (scaling raw wave signals into a range of [-1, 1]) per utterance, followed by pre-emphasis (0.97), and extracted the spectrogram  with STFT (50-ms frame length, 12.5-ms frame shift, 2048-point FFT). After getting the spectrogram, we used the squared magnitude and a Mel-scale filterbank with 40 filters to extract the Mel-scale spectrogram. After getting the Mel-spectrogram, we squared  the magnitude spectrogram features. In the end, we transformed each speech utterance into a log-scale and normalized each feature into 0 mean and unit variances. Our final set is comprised of 40 dims  log Mel-spectrogram features and a 1025 dims log magnitude spectrogram.
	
	For the text, we converted all of the sentences into lowercase and replaced some punctuation marks (for example, " into '). In the end, we have 26 letters (a-z), six punctuation marks (,:'?.-), and three special tags ($<$s$>$, $<$/s$>$, $<$spc$>$) to denote start, end of sentence, and spaces between words.

	\subsection{Model Details}
	 
	Our ASR model is a standard encoder-decoder with an attention mechanism. On the encoder side, we used a log-Mel spectrogram as the input features (in unsupervised process, the log Mel-spectrogram was generated by TTS), which are projected by a fully connected layer and a LeakyReLU ($l=1e-2$) \cite{xu2015empirical} activation function and processed by three stacked BiLSTM layers with 256 hidden units for each direction (512 hidden units). We applied sequence subsampling \cite{cho2014learning, chan2016listen} on the top two layers and reduced the length of the speech features by a factor of 4. On the decoder side, the input character is projected with a 128 dims embedding layer and fed into a one-layer  LSTM with 512 hidden units. We calculated the attention matrix with an MLP scorer (Eq. \ref{eq:mlpscore}), followed by a fully connected layer and a softmax function. Both the ASR and TTS models are implemented with the PyTorch library \footnote{PyTorch \url{https://github.com/pytorch/pytorch}}.
	
	Our TTS model hyperparameters are generally the same as the original Tacotron, except that we used LeakyReLU instead of ReLU for most of the parts. On the encoder sides, the CBHG used $K=8$ different filter banks instead of 16 to reduce our GPU memory consumption. For the decoder sides, we used a two-stacked LSTM instead of a GRU with 256 hidden units. Our TTS predicted four consecutive frames in one time step to reduce the number of time steps in the decoding process.

	\subsection{Experiment Result}
	 
	Table \ref{tbl:single_spk} shows our result on the single-speaker ASR and TTS experiments. For the ASR experiment, we generated best hypothesis with beam search (size$=5$). We used a character error rate (CER) for evaluating the ASR model. For the TTS experiment, we reported the MSE between the predicted log Mel and the log magnitude spectrogram to the ground truth. We also report the accuracy of our model that predicted the last speech frame.  We used different values for $\alpha$ and text decoding strategy for ASR (in the unsupervised learning stage) with a greedy search or a beam search.

	\begin{table}[h]
		\centering
		\caption{Experiment result for single-speaker test set.}
		\label{tbl:single_spk}\footnotesize
		\begin{tabular}{|c|c|c|c|c|c|c|c|}
			\hline
			\multirow{2}{*}{\textbf{Data}}                                               & \multicolumn{3}{c|}{\textbf{Hyperparameters}}                                                 & \textbf{ASR}                                                & \multicolumn{3}{c|}{\textbf{TTS}}                                                         \\ \cline{2-8}
			& $\alpha$ & $\beta$ & \textbf{\begin{tabular}[c]{@{}c@{}}gen.\\ mode\end{tabular}} & \textbf{\begin{tabular}[c]{@{}c@{}}CER\\ (\%)\end{tabular}} & \textbf{Mel} & \textbf{Raw} & \textbf{\begin{tabular}[c]{@{}c@{}}Acc\\ (\%)\end{tabular}} \\ \hline
			\begin{tabular}[c]{@{}c@{}}Paired \\ (10k)\end{tabular}                      & -              & -             & -                                                            & 10.06                                                       & 7.068        & 9.376        & 97.7                                                        \\ \hline
			\multirow{4}{*}{\begin{tabular}[c]{@{}c@{}}+ Unpaired \\ (40k)\end{tabular}} & 0.25           & 1             & greedy                                                       & 5.83                                                        & 6.212        & 8.485        & 98.4                                                        \\ \cline{2-8}
			& 0.5            & 1             & greedy                                                       & 5.75                                                        & 6.247        & 8.418        & 98.4                                                        \\ \cline{2-8}
			& 0.25           & 1             & beam 5                                                       & 5.44                                                        & 6.243        & 8.441        & 98.3                                                        \\ \cline{2-8}
			& 0.5            & 1             & beam 5                                                       & 5.77                                                        & 6.201        & 8.435        & 98.3                                                        \\ \hline
		\end{tabular}
	\end{table}

	The result show that after ASR and TTS models have been trained with a small paired dataset, they start to teach each other using unpaired data and generate useful feedback. Here we improved both ASR and TTS performance. Our ASR model reduced CER by 4.6\% compared to the system that was only trained with labeled data. In addition to ASR, our TTS also decreased the MSE and the end of speech prediction accuracy.

	\section{Experiment on Multi-Speaker Task}
	 
	Based on our good result on a single speaker, we extended our initial work to a multi-speaker experiment. Unlike our previous experiment, we used a real natural speech corpus instead of synthesized speech. We used the BTEC ATR-EDB \cite{btecedb2012} corpus, which contains about 180,000 speech utterances from six different regions (Australia, British, US West, US Northeast, US South, and US West ). In this experiment, we only used the US Northeast portion in the dataset (contains about 22000 utterances). The US Northeast contains about 50 different speakers (25 males, 25 females) and about 440 utterances per speaker. For training, validation, and testing sets, we split the dataset by 20 utterances per speaker for validation, 20 utterances per speaker for testing, and the rest for training. For the paired speech and text, we got 80 pairs per speaker and the rest of the speech and text were used as unsupervised training sets. None of these training sets (paired and unpaired) or the validation and test sets overlap.

	\subsection{Model Details}
	 
	In this experiment, we used the same ASR model as in the previous section without any modifications. However, for the TTS, we used the modified Tacotron with speaker embedding. Here, we used the pre-trained model from a single speaker and transferred the weight except for the speaker-embedding layer.

	\subsection{Experiment Result}
	 
	% \usepackage{multirow}
	
	\begin{table}[h]
		\centering
		\caption{Experiment result for multi-speaker test set.}
		\footnotesize
		\label{tbl:multi_spk}
		\begin{tabular}{|c|c|c|c|c|c|c|c|}
			\hline
			\multirow{2}{*}{\textbf{Data}}                                                     & \multicolumn{3}{c|}{\textbf{Hyperparameters}}                                                 & \textbf{ASR}                                                & \multicolumn{3}{c|}{\textbf{TTS}}                                                         \\ \cline{2-8}
			& $\alpha$ & $\beta$ & \textbf{\begin{tabular}[c]{@{}c@{}}gen.\\ mode\end{tabular}} & \textbf{\begin{tabular}[c]{@{}c@{}}CER\\ (\%)\end{tabular}} & \textbf{Mel} & \textbf{Raw} & \textbf{\begin{tabular}[c]{@{}c@{}}Acc\\ (\%)\end{tabular}} \\ \hline
			\begin{tabular}[c]{@{}c@{}}Paired \\ (80 utt/spk)\end{tabular}                     & -              & -             & -                                                            & 26.47                                                       & 10.213       & 13.175       & 98.6                                                        \\ \hline
			\multirow{4}{*}{\begin{tabular}[c]{@{}c@{}}+ Unpaired \\ (remaining)\end{tabular}} & 0.25           & 1             & greedy                                                       & 23.03                                                       & 9.137        & 12.863       & 98.7                                                        \\ \cline{2-8}
			& 0.5            & 1             & greedy                                                       & 20.91                                                       & 9.312        & 12.882       & 98.6                                                        \\ \cline{2-8}
			& 0.25           & 1             & beam 5                                                       & 22.55                                                       & 9.359        & 12.767       & 98.6                                                        \\ \cline{2-8}
			& 0.5            & 1             & beam 5                                                       & 19.99                                                       & 9.198        & 12.839       & 98.6                                                        \\ \hline
		\end{tabular}
	\end{table}

	Table \ref{tbl:multi_spk} reports our result on the multi-speaker ASR and TTS experiments. Similar with the single speaker result, we improved both the ASR and TTS models by additional training on unpaired datasets. However, in this experiment, we found that a different ASR performance $\alpha=0.5$ produced a larger improvement than $\alpha=0.25$. We hypothesize that because the baseline model is not as good as the previous single speaker experiment, we should put a larger coefficient on the loss and the gradient provided by the paired training set.

	\section{Related Works}
	
	Approaches that utilize learning from source-to-target and vice-versa as well as feedback links remain scant. He et al. \cite{he2016dual} quite recently published a work that addressed a mechanism called dual learning in neural machine translation. Their system has a dual task: source-to-target language translation (primal) versus target-to-source language translation (dual). The primal and dual tasks form a closed loop and generate informative feedback signals to train the translation models, even without the involvement of a human labeler. 
	This approach was originally proposed to tackle training data bottleneck problems. With a dual-learning mechanism, the system can leverage monolingual data (in both the source and target languages) more effectively. First, they construct one model to translate from the source to the target language and another to translate from the target to the source language. After both the first and second models have been trained with a small parallel corpus, they start to teach each other using monolingual data and generate useful feedback with language model likelihood and reconstruction error to further improve the performance.
	
	Another similar work in neural machine translation was introduced by Cheng et al. \cite{cheng2016semi}. This approach also exploited monolingual corpora to improve neural machine translation. Their system utilizes a semi-supervised  approach  for  training  neural machine translation (NMT) models  on  the  concatenation  of  labeled (parallel  corpora)  and  unlabeled  (mono-lingual corpora) data.  The central idea is to reconstruct monolingual corpora using an autoencoder in which the source-to-target and  target-to-source translation models serve as the encoder and decoder, respectively.
	
	However, no studies have yet addressed similar problems in spoken language processing tasks. This paper presents a novel mechanism 
	that integrates human speech perception and production behaviors. With a concept that resembles dual learning in neural machine translation, we utilize the primal model (ASR) that transcribes the text given the speech versus the dual model (TTS) that synthesizes the speech given the text. However, the main difference between NMT  is that the domain between the source and the target here are different (speech versus text). While ASR transcribes the unlabeled speech features, TTS attempts to reconstruct the original speech waveform based on the text from ASR. In the opposite direction, ASR also attempts to reconstruct the original text transcription given the synthesized speech. Nevertheless, our experimental results show that the proposed approach also identified a successful learning strategy and significantly improved the performance more than separate systems that were only trained with labeled data.

	\section{Conclusion}
	
	This paper demonstrated a novel machine speech chain mechanism
	based on deep learning. The sequence-to-sequence
	model in closed-loop architecture allows us to train our model
	on the concatenation of both labeled and unlabeled data. We
	explored applications of the model in various tasks, including
	single speaker synthetic speech and multi-speaker natural
	speech. Our experimental results in both cases show that the
	proposed approach enabled ASR and TTS to further improved
	the performance by teaching each other using only unpaired data.
	In the future, it is necessary to further validate
	the effectiveness of our approach on various languages and
	conditions (i.e., spontaneous, noisy, and emotion). 
	
	\section{Acknowledgements}
	
	Part of this work was supported by JSPS KAKENHI Grant Numbers JP17H06101 and JP 17K00237.
	
	\bibliographystyle{IEEEtran}
	\bibliography{mybib}
	
\end{document}